\documentclass[]{article}
\usepackage[pdftex]{graphicx}

\begin{document}

\title{An exploration of asocial and social learning in the evolution of variable-length structures}

\author{{Michael O'Neill and Anthony Brabazon}\\
Natural Computing Research \& Applications Group\\School of Business\\
University College Dublin\\
Ireland\\
Email: m.oneill@ucd.ie | anthony.brabazon@ucd.ie}

\maketitle

\begin{abstract}
We wish to explore the contribution that asocial and social learning might play as a mechanism for self-adaptation in the search for variable-length structures by an evolutionary algorithm. An extremely challenging, yet simple to understand problem landscape is adopted where the probability of randomly finding a solution is approximately one in a trillion. A number of learning mechanisms operating on variable-length structures are implemented and their performance analysed. The social learning setup, which combines forms of both social and asocial learning in combination with evolution is found to be most performant, while the setups exclusively adopting evolution are incapable of finding solutions.
\end{abstract}

\section{Introduction}
Hinton \& Nowlan~\cite{hinton&nowlan} demonstrated potential benefits of lifetime learning for evolution in the form of a Baldwin effect~\cite{baldwin}. They examined a problem domain that required a predefined and fixed number of phenotypic traits, which were encoded in a fixed-length genotype. The landscape was a challenging one for evolution as it took the form of a needle-in-a-haystack, with no gradient available to guide the population towards the ideal target. With the inclusion of a learning mechanism additional information became available to the population, and that resulted in the combination of evolution and learning being capable of finding solutions. Evolution alone was unable to find a solution given the population size employed. 

In this study we adapt the Hinton \& Nowlan problem landscape to explore the utility of learning in the space of variable-length phenotypes where the number of required traits is not predefined and can vary between individuals within a population, such as exists within genetic programming populations. The move from a fixed-length to variable-length problem, where the space of structures must be explored in addition to optimising the content, increases the search space by orders of magnitude rendering the problem significantly more challenging. We compare a number of approaches to learning including a number of variants of individual, or asocial learning, and a social learning strategy, to understand their behaviour in problems of a variable-length nature.

The following section provides some background and context to the study focusing on the interplay between learning and evolution. Section~\ref{expts} details the experimental setup where we set out to explore the contribution that learning might play in the search of the space of variable-length structures. Results and Conclusions are presented in Sections~\ref{results} and~\ref{conc} respectively.

\section{Background}

There are two main categories of learning in the natural world, namely social and asocial learning~\cite{heyes}. Asocial learning involves a change in an organisms behaviour in response to a specific experience, which might include a stimulus such as an event or object in the environment, a response to a stimulus, or some relationship between two stimuli. On the other hand social learning involves organisms imitating or transmitting  behaviour to one another. Learning then is operating at the level of the phenotype.

There have been many studies of learning and evolution in the evolutionary computation and artificial life (e.g., ~\cite{ackley},~\cite{nolfi},~\cite{bull},~\cite{nam}) literature's, too many to detail here, as such a sample of the seminal work relevant to this study is captured. It is worth highlighting that until recently the majority of studies have exclusively involved asocial learning, and few if any have examined social learning in the search of variable-length structures.

\begin{figure*}[!t]
\centering
\includegraphics[width=\textwidth]{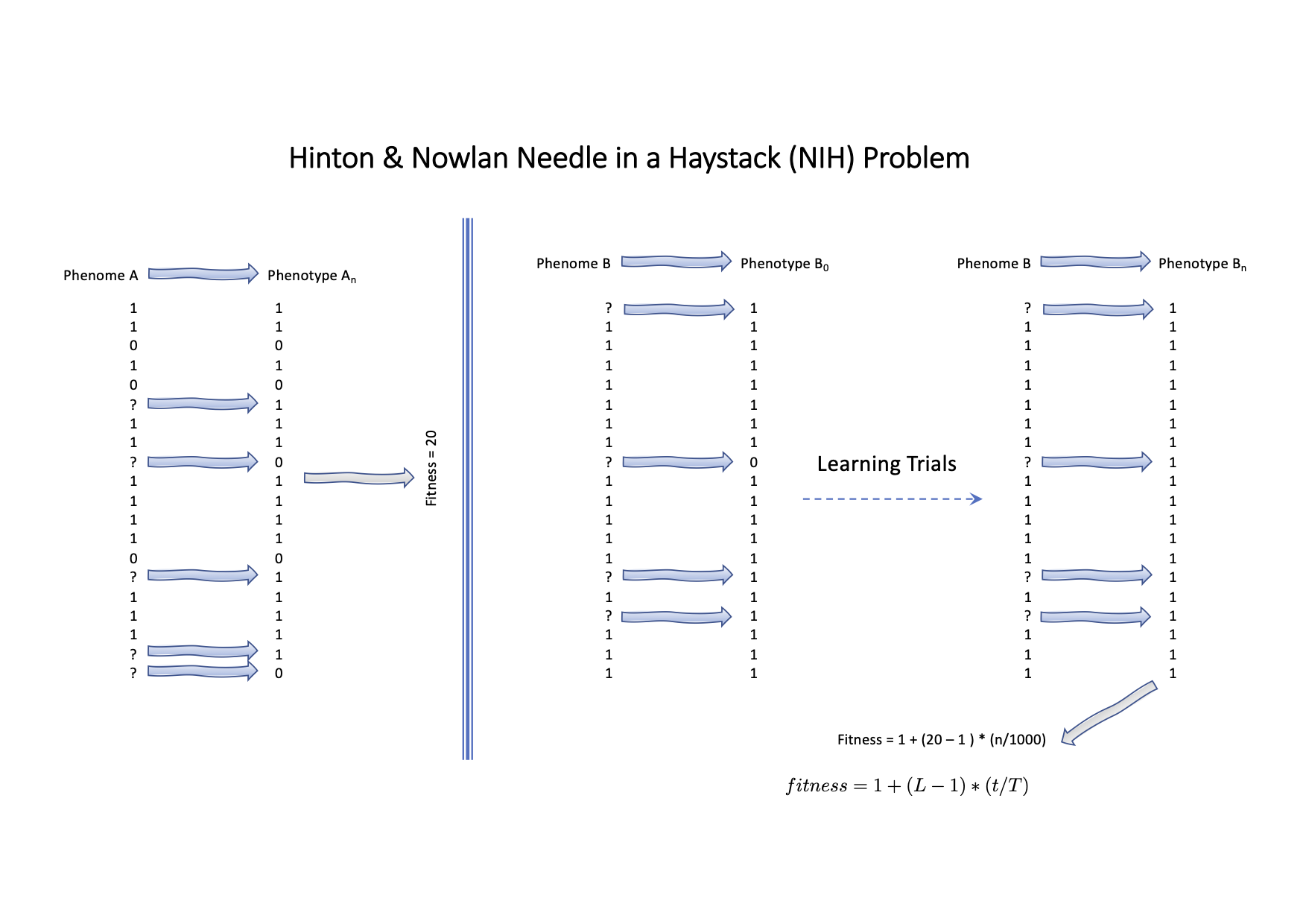}
\caption{An illustration of the Hinton \& Nowlan fitness landscape. Phenomes containing a 0 at any locus automatically receive the worst possible fitness of 20. Phenome A on the left-hand-side of the figure illustrates this scenario. On the right-hand-side of the figure we see each phenome (illustrated here using Phenome B) is given a number of learning trials (1000) to generate phenotypes (Phenotype$_0$ through to Phenotype$_n$ where the maximum n in this study is 1000) from the phenome by converting the plastic symbols to either a 0 or 1. If one of the phenotypes generated is comprised of all 1's (i.e., the perfect solution) the fitness of the phenome is calculated based on the number of learning trials taken to find that perfect solution. The fewer trials taken the better the fitness of the phenome. }
\label{HNNIH}
\end{figure*}

Hinton \& Nowlan~\cite{hinton&nowlan} explore the interaction between learning and evolution with the adoption of a primitive learning mechanism. The phenotype of each individual in the population is specified by a genome with twenty loci. Each locus can take on one of three symbols (0, 1 or ?). The ? symbol is said to be {\it plastic} in that it represents either a 0 or 1. To evaluate the fitness of an individual each plastic symbol must be resolved to either a 0 or 1. As such, the presence of a ? means there is potential for that individual to learn during their lifetime what better values of ? might be in the current environment. The fitness landscape is straightforward (see Fig.~\ref{HNNIH} for an illustration). If an individual contains a 0, it is awarded the worst possible fitness. If an individual contains all 1's, it is rewarded with the best fitness. If an individual contains both 1's and ?'s, if all the ?'s are resolved to become 1's then the best possible fitness is adjusted by reducing the fitness in proportion to the number of learning trials that were undertaken.

Learning and evolution are both then forms of adaptation. Mayley~\cite{mayley} discusses the benefits and costs that lifetime adaptation through learning brings to a population undertaking simulations on a tunably rugged NK landscape~\cite{kauffman}.

\begin{figure*}[!hbt]
\centering
%Nolearning setup grammar
\footnotesize
\begin{verbatim}
<phenome> ::= <s><s><s><s><s><s><s><s><s><s><s><s><s><s><s><s><s><s><s><s>
<s> ::= <zero>|<one>
<one> ::= 1
<zero> ::= 0
\end{verbatim}
\caption{nolearning grammar for the classic fixed-length problem with twenty terminal symbols (1's or 0's) in the phenotype.}
\label{FixedLenGrammar}
\end{figure*}

\begin{figure*}[!hbt]
\centering
\footnotesize
\begin{verbatim}
<phenome> ::= <s><s><s><s><s><s><s><s><s><s><s><s><s><s><s><s><s><s><s><s>
<s> ::= <zero>|<one>|?|?
<one> ::= 1
<zero> ::= 0
\end{verbatim}
\caption{The grammar (asocial learning setup) for the fixed-length problem with twenty terminal symbols (1's or 0's) in the phenotype. This grammar includes the ability of the phenome to include plastic ? symbols, which are resolved to one of the terminal symbols (1 or 0) during learning. Note as per the original experimental setup reported in~\cite{hinton&nowlan} the grammar enables that on average 50\% of an individual will become a plastic symbol during intialisation of the first generation. Correspondingly, on average 25\% of an individual with either be a 0 or a 1.}
\label{FixedLenGrammarWithPlasticSymbols}
\end{figure*}

Gruau and Whitley~\cite{gruau} explore asocial learning with Cellular Encoding with both Baldwinian~\cite{baldwin} and Lamarckian~\cite{lamarck} mechanisms. Social learning is not considered. The purely Baldwinian setups allow learning to modify the weights of the evolved neural networks. The Lamarckian setups allow the information that is learned to be encoded in the genome and are so afforded the opportunity to pass on these acquired characteristics through the process of evolution.
In this study we restrict ourselves to the classic separation of genotype and phenotype according to the central dogma of molecular biology, that information flows from genotype to phenotype (and not backwards), such that learning takes place through modifications to  the phenotype, which are not then encoded back on the genome. Evolution then operates exclusively on the genome. In recent years what might be considered violations of the central dogma of molecular biology have been observed in developmental biology, for example, in the form of genomic imprinting~\cite{imprinting} where epigenetic modifications can change the expression of genes. An elaborate developmental process~\cite{ncabook} is not adopted here and so we exclude the possibility of transmission of epigenetic features.

In this study we set out to explore the contribution that learning might play as a mechanism for self-adaptation in the search for variable-length structures, such as exists within genetic programming. A novel aspect of this study is that we consider both asocial and social learning in the presence of variable-length structures. To this end we examine a very challenging, yet simple to understand, problem landscape inspired by the seminal study of Hinton \& Nowlan~\cite{hinton&nowlan}, which demonstrated the adaptive advantage that a learning mechanism can bring to an evolutionary system. The sheer scale of the increased search space size adopted here coupled to the relatively small population size and number of learning trials mitigates previous criticisms~\cite{belew} of the classic Hinton \& Nowlan~\cite{hinton&nowlan} setup.

\begin{figure*}[!t]
\centering
\includegraphics[width=\textwidth]{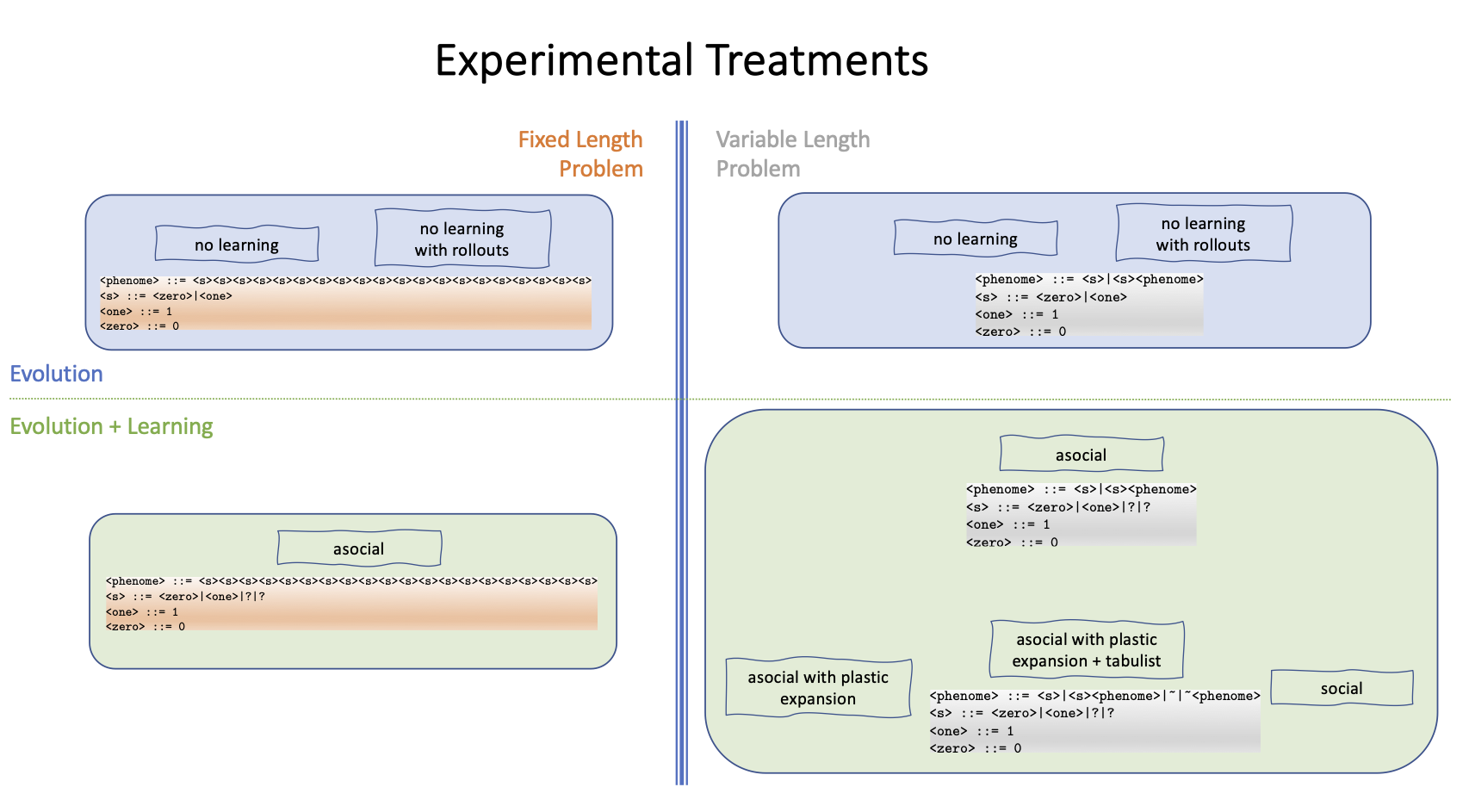}
\caption{An overview of the experimental treatments adopted in this study. Control treatments include the no learning setups, which are restricted to evolutionary search both with and without a form of local search that generates a number of random variants of the phenotype. The number of random variants is equivalent to the number of Learning Trials (i.e., 1000) adopted in the treatments that include both evolution and learning. The control setups therefore explore the maximum number of phenotypes that the learning setups are capable of exploring. We then have treatments exclusively with asocial learning where plastic symbols are resolved randomly, or alternatively using a tabulist which represents a memory of previously visited phenotypes. The final setup adopts a form of social learning in which individuals share their phenotype lengths and use this information to adjust the lengths of their individual phenotype by controlling the number of expansions of their structural plastic symbols.}
\label{expttreatments}
\end{figure*}

\section{Experimental Setup}
\label{expts}
\begin{figure*}[!t]
\centering
\includegraphics[width=\textwidth]{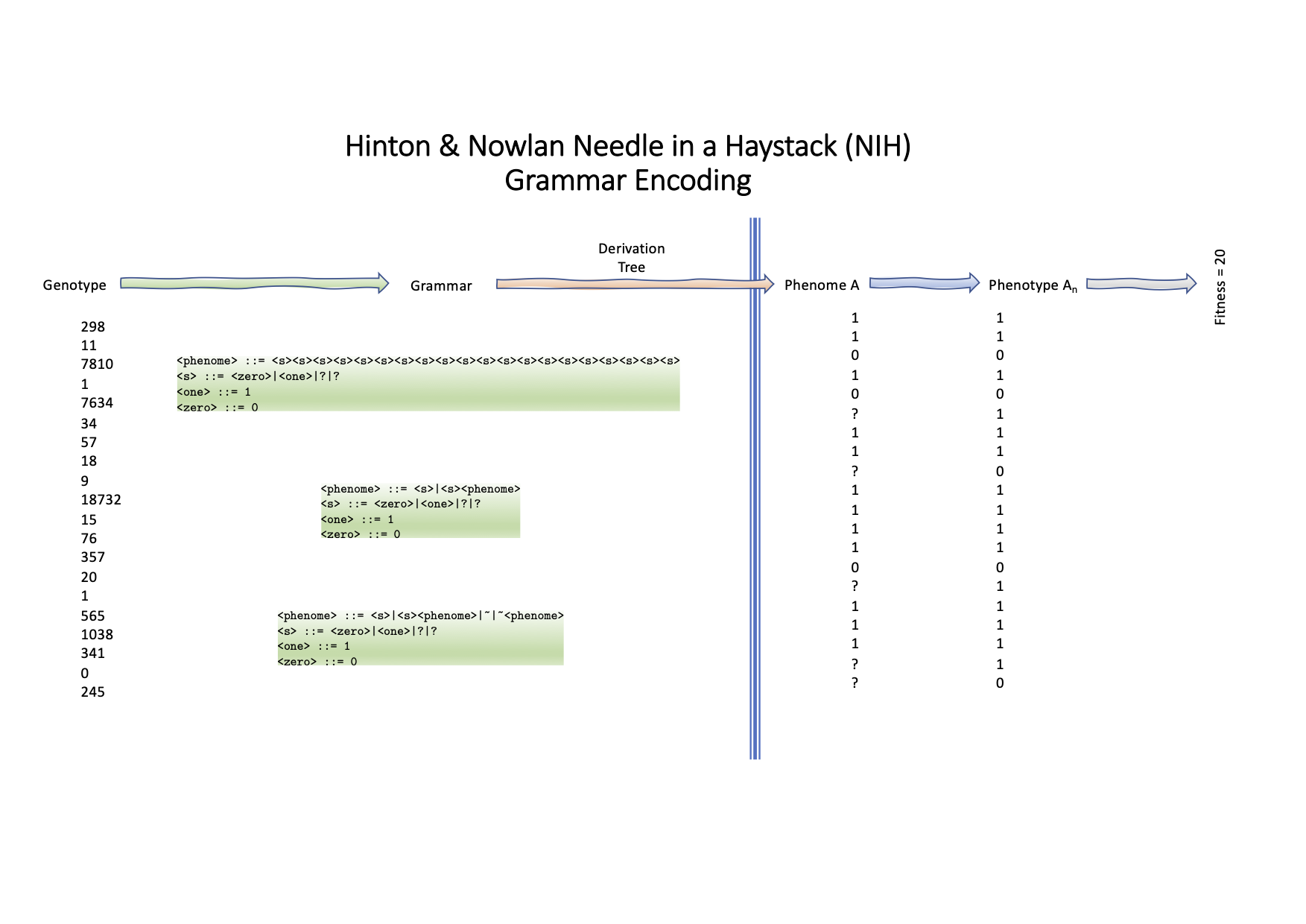}
\caption{An illustration of the mapping from genotype through a grammar onto a derivation tree, which is resolved to the phenome before learning is permitted to generate phenotypes. }
\label{HNNIHgrammarmap}
\end{figure*}

An overview of the experimental treatments adopted in this study is presented in Fig.~\ref{expttreatments}. We wish to explore the contribution that learning (both asocial and social) might play in the search for variable-length structures. To this end a python implementation of grammar-based genetic programming~\cite{mckay2010} using PonyGE2~\cite{ponyge2} is adopted in this study. For each learning variant we adopt a linear genotype encoding search operator, Grammatical Evolution type algorithm. An illustration of the map from genotype to phenotype is presented in Fig.~\ref{HNNIHgrammarmap}. In each case the so-called PIGrow initialisation method from PonyGE2 is used with the same tree-depth initialisation restrictions. 100 replications of each algorithm are run with the same set of pseudo random number generator seeds. The parameter settings are detailed in Table~\ref{parameters}.

\begin{table}[!hbt]
\renewcommand{\arraystretch}{1.3}
\caption{The parameter settings adopted for each experimental setup.}
\label{parameters}
\centering
\begin{tabular}{rl}
{\bf Parameter} & {\bf Setting} \\
Population Size & 1,000\\
Generations & 50\\
Mutation & integer (1 event per individual)\\
Crossover & variable\_onepoint\\
Crossover Probability & 0.9\\
Selection & Tournament (size=2)\\
Replacement & generational with elitism (10 individuals)\\
Max Initialisation Tree Depth & 10\\
Max Tree Depth & 50\\
Initialisation & PIGrow\\
Learning Trials & 1,000\\
Target Solution& 11111111111111111111 \\
\end{tabular}
\end{table}

\begin{figure*}[!hbt]
\centering
\begin{verbatim}
<phenome> ::= <s>|<s><phenome>
<s> ::= <zero>|<one>|?|?
<one> ::= 1
<zero> ::= 0
\end{verbatim}
\caption{The grammar for the variable-length problem shown with plastic symbols. The grammar for the nolearning setup on this problem simply removes the plastic (?) rules from the grammar.}
\label{VariableLenGrammar}
\end{figure*}

\begin{figure*}[!hbt]
\centering
\begin{verbatim}
<phenome> ::= <s>|<s><phenome>|~|~<phenome>
<s> ::= <zero>|<one>|?|?
<one> ::= 1
<zero> ::= 0
\end{verbatim}
\caption{The grammar for the variable-length problem shown with the inclusion of plastic expansion symbols in the {\tt phenome} rule. Note that \textasciitilde{} is equivalent to a ? plastic symbol with the exception that it is rooted in the context of $<$phenome$>$ (a structural rule) rather than $<$s$>$ (a content rule). We adopt the separate \textasciitilde{} symbol to note this typing distinction so that we can later easily observe which plastic symbols are responsible for structural search versus those responsible for content search (i.e., choosing between 1's and 0's) search.}
\label{VariableLenPlasticExpansionGrammar}
\end{figure*}

In the first experiments we set out to observe the behaviour and performance of the grammar-based genetic programming algorithms on the classic Hinton \& Nowlan fixed-length problem~\cite{hinton&nowlan} with a target phenotype vector of twenty 1's. Three learning variants are explored. The first two are control benchmarks where the algorithms exclusively adopt an evolutionary search with no learning. There are no plastic (?) symbols available to be encoded, and the grammar is presented in Figure~\ref{FixedLenGrammar}. The first control setup ({\bf nolearning}) simply adopts a population size of 1,000 running for 50 generations. The second control ({\bf nolearning with rollouts}) extends the nolearning setup by allowing a number of variants of each phenotype to be explored in a kind of random local search. The number of variants generated is equivalent to the number of learning trials adopted in the learning setups. The number of learning trials is a parameter setting which is set to 1,000 across all the experiments conducted in this study, with this value equivalent to that adopted in ~\cite{hinton&nowlan}. The third setup allows plastic (?) symbols to be expressed in the phenome with the grammar detailed in Figure~\ref{FixedLenGrammarWithPlasticSymbols}. In the learning setup ({\bf asocial}) phenomes that contain plastic symbols must replace each ? randomly with either a '1' or a '0' in order to resolve to a phenotype that can be evaluated. Please note that we adopt the term phenome to represent an incompletely mapped solution that contains plastic symbols. A phenotype refers to the completely mapped vector comprised exclusively of terminal symbols (i.e., \verb*z1z's and \verb*y0y's). Learning takes place by allowing 1,000 learning trials to occur, where up to 1,000 variants of the phenome are expressed (i.e., replacements of ? plastic symbols). 

In the second experiment we explore a number of variants of learning on a variable-length version of the classic Hinton \& Nowlan problem~\cite{hinton&nowlan}. Solutions to this problem must successfully encode the number of features in the phenotype vector, in addition to the symbol for each feature. The grammars for this problem are detailed in Figure~\ref{VariableLenGrammar}. While the probability of solving the fixed-length version of the problem by chance is 1 in $2^{20}$ (approximately 1 in a million), for the variable-length version of the problem this reduces sharply to $2^{40}$ (approximately 1 in a trillion). Consequently this variable-length form of the needle-in-a-haystack problem is extremely challenging for the relatively small population sizes (1,000) and learning trials (1,000) adopted in this study.

Taking into consideration the variable-length nature of the problem landscape where simultaneously both the solution structure and its constituent symbolic values need to be explored, three additional learning setups are adopted ({\bf plastic expansion}, {\bf tabulist} and {\bf social}). 

The first ({\bf plastic expansion}) adopts asocial learning and adds a new plastic symbol \verb*z~z to the expansion rule \verb*z<phenome>z of the grammar (see Figure~\ref{VariableLenPlasticExpansionGrammar}). Note that \verb*z~z is equivalent to a ? plastic symbol with the exception that it is rooted in the context of \verb*z<phenome>z rather than \verb*z<s>z. We adopt the separate \verb*z~z symbol for convenience to highlight this typing distinction so that we can later easily observe which plastic symbols are responsible for structural search versus those responsible for content search (i.e., choosing between \verb*z1z's and \verb*z0z's) search. As per the original plastic symbol ?, each \verb*z~z that appears in the phenome must be resolved to a terminal symbol (i.e., one of \verb*z1z or \verb*z0z). During the mapping of the genotype to phenotype each \verb*z~z is randomly replaced (in accordance with the grammar to preserve syntactic correctness of the phenotype) with either a ? or a \verb*z~z? until all \verb*z~z's are resolved in the first instance to a ?. The learning mechanism is, therefore, capable of expanding the length of the expressed phenotype. Then each ? is resolved to one of \verb*z1z or \verb*z0z randomly (as per the earlier asocial learning setups). A predefined maximum number of learning trials (or variants) of the phenome are expressed when plastic symbols are present, as per the earlier asocial learning setup. The number of learning trials is again 1,000. 

If during the generation of phenotype variants during the learning trial phase, a perfect solution occurs the learning terminates and a fitness is returned. For both this and the earlier asocial learning setup the fitness is calculated according to~\cite{hinton&nowlan} as:

\begin{equation}
    fitness = 1 + (L-1) * (t / T)
\end{equation}

where $L$ is the length of the ideal target (20 in this study), $t$ is the
number of learning trials taken to find the target, and $T$ is the maximum 
number of learning trials available (1,000 across all experimental setups).
When the ideal target is not found the fitness of the individual is set to the maximum (worst) possible (i.e., L=20). All experimental setups are therefore attempting to minimise the fitness value to be as close to zero as possible.

The second alternative learning setup ({\bf tabulist}) employs a tabulist from Tabu Search~\cite{tabu}, such that, during learning trials where we randomly rollout the plastic symbols to terminals (\verb*z1z's or \verb*z0z's) that if we create a phenotype that has been found before by that individual (i.e., it exists in the tabulist) we attempt to generate a new phenotype by re-replacing the plastic symbols. Given it will not always be possible to generate a novel phenotype from any phenome (due to the number of plastic symbols in the phenome) we restrict the tabulist checks to a maximum limit of 10. From a learning perspective we have provided each individual with a memory of previously visited/expressed phenotypes over the duration of their "lifetime" (i.e., one generation's worth of learning trials).

The third additional setup ({\bf social}) adopts a simple form of social learning. Each individual shares with the population the length of the largest phenotype it has generated during its lifetime (i.e., during a single generation of the evolutionary algorithm). During the mapping from the genotype to phenotype every individual in the population can "see" the length of the phenotype of the best fitness solution found to date. This population maximum is taken into consideration when mapping the plastic expansion symbols (\verb*z~z's), such that each \verb*z~z in the individual is expanded to a number of \verb*z?z's according to:

\begin{equation}
    PL^{'} = PL + | MPL - IPL |
\end{equation}

where $PL$ is the length of the phenome, MPL is the maximum phenotype length of the best individual in the population (the social learning term), and IPL is the maximum phenotype length expressed by this individuals phenome to date during its learning trials (an asocial learning term, or personal information). $PL^{'}$ is then the length of the phenome with all \verb*z~z's resolved to \verb*z?z's.

Please note that during the instantiation of the population of the first generation the length of the phenotype of the best solution is not yet known. As such during this phase mapping of the genotype to the pheneotype follows the earlier asocial setup, with social learning taking place during each subsequent generation.

\begin{figure*}[!t]
\centering
\includegraphics[width=4.5in]{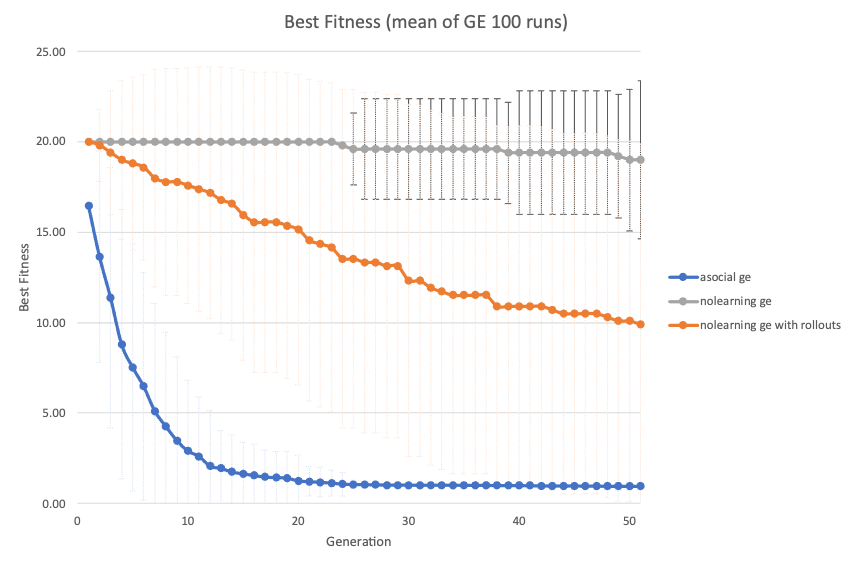}
\caption{Mean best fitness performance on Hinton \& Nowlan needle-in-a-haystack landscape for their classic fixed-length encoding representing a phenotypic vector with twenty traits.}
\label{HNFixedLenFitness}
\end{figure*}

\begin{figure*}[!t]
\centering
\includegraphics[width=\textwidth]{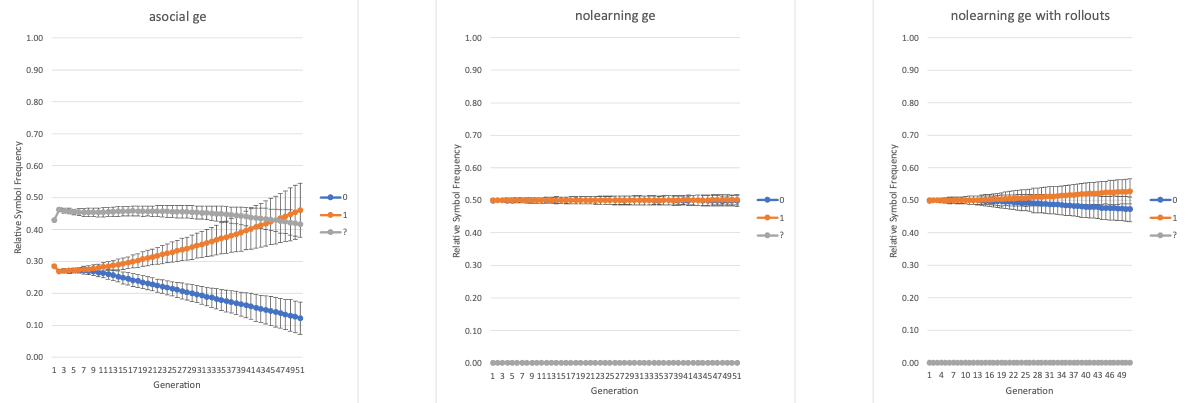}
\caption{Phenome terminal symbol relative frequency (1, 0 and the plastic symbol ?) on Hinton \& Nowlan needle-in-a-haystack landscape for their classic fixed-length encoding representing a phenotypic vector with twenty traits (the ideal target is twenty 1's).}
\label{HNFixedLenSymbols}
\end{figure*}

\begin{figure*}[!t]
\centering
\includegraphics[width=4.5in]{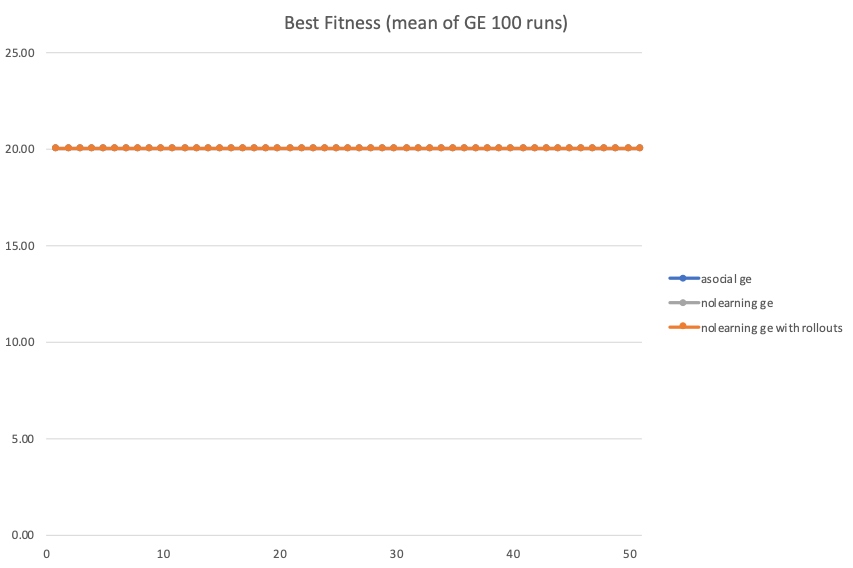}
\caption{Mean best fitness performance on Hinton \& Nowlan needle-in-a-haystack landscape for the variable-length encoding representing a phenotypic vector with twenty traits. The same algorithms from the classic fixed-length version of the problem fail to find correct solutions.}
\label{HNVarLenFitnessOriginal}

\end{figure*}
\begin{figure*}[!t]
\centering
\includegraphics[width=4.5in]{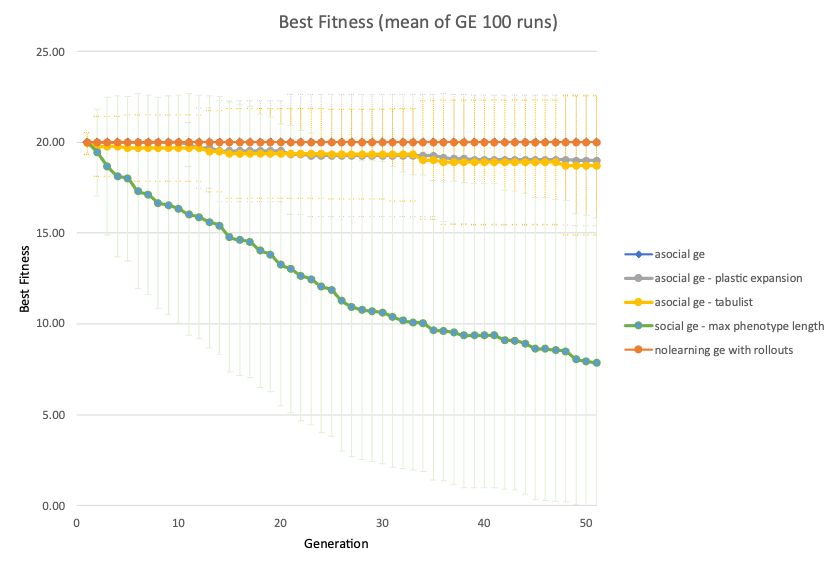}
\caption{Mean best fitness performance on Hinton \& Nowlan needle-in-a-haystack landscape for different learning variants examined on the variable-length encoding representing a phenotypic vector with twenty traits. The social learning setup that shares the maximum phenotype length found by the best solution to date is observed to have better performance.}
\label{HNVarLenFitnessVariants}
\end{figure*}

\section{Results}
\label{results}
Results for the classic fixed-length problem instance are captured in Figures~\ref{HNFixedLenFitness} and \ref{HNFixedLenSymbols}. We observe similar behaviour to that found in ~\cite{hinton&nowlan}, with the pure evolutionary setup ({\bf nolearning}) unable to consistently find correct solutions. When we allow a random local search without learning ({\bf no learning with rollouts}) some replications are more likely to find solutions. This experimental setup was not used in~\cite{hinton&nowlan}, however, we believe it is a fairer control as it is afforded the opportuntity to explore the same number of phenotypes that are available to the setups with learning. Effectively the sample size made by the population over the course of a run in the {\bf no learning with rollouts} setup and the {\bf asocial} setups are equivalent. Differences observed in behaviour of these two setups are less likely to be due to the ability to randomly find a solution. In the setup with learning ({\bf asocial ge}) solutions are consistently found, and are found in much earlier generations than the other setups. In terms of the proportion of symbols in the phenome we again observe similar trends to those in ~\cite{hinton&nowlan} with a preservation of the plastic symbol ? and a corresponding growth and decline in \verb*z1z's and \verb*z0z's respectively. Recall the ideal target is comprised of twenty \verb*z1z's. 

Turning to the much more challenging, variable-length instance of the problem, when we examine the behaviours on the same setups adopted on the fixed-length problem instance, we observe that no setup is capable of finding a solution (see Figure~\ref{HNVarLenFitnessOriginal}). Given the significant increase in the size of the search space when both the structure/length of the solution must be found, in addition to the correct symbols at each phenotype locus, the learning mechanisms in these setups is insufficient. 

Results for the three learning variants ({\bf plastic expansion}, {\bf tabulist}, and {\bf social} learning) are provided in Figure~\ref{HNVarLenFitnessVariants} where we observe the social learning setup in particular being capable of finding solutions to the problem. Both the plastic expansion and tabulist (which includes plastic expansion) are observed to have similar behaviour and unlike the earlier setups some populations do find a solution. 

The number of replications out of 100 performed for each setup where a successful solution is found is detailed in Table~\ref{successfulruns}. It is clear that when learning is capable of exploring the length of phenotype solutions and simultaneously capable of taking into consideration the length of the best solution in the population (combining both social and asocial learning) that together with evolution solutions to this challenging problem landscape can be successfully found.

Examining the mean lengths of phenotypes generated by each setup (see Table~\ref{phenotypelengthstable}), we observe that the more successful setups are producing longer phenotypes, and the lengths of those phenotypes are on average close to the  length of the target phenotype. The less successful setups appear to be oversampling smaller solutions, which is a common issue for Genetic Programming and well supported by theory and perhaps counter intuitively is the best explanation to date for the existence of bloat~\cite{poli:langdon:dignum:2007}~\cite{dignum:poli:2007}. The relatively simple social learning setup appears to successfully overcome this size sampling limitation allowing a more effective search of the space of variable-length structures.

\begin{table*}[!t]
\renewcommand{\arraystretch}{1.3}
\caption{The number of replications out of 100 on the variable-length version of the Hinton \& Nowlan needle-in-a-haystack landscape that find a correct phenotype solution.}
\label{successfulruns}
\centering
\footnotesize
\begin{tabular}{cccccc}
&    nolearning  &  &      asocial &   asocial&        social   \\
&    with rollouts &  asocial&      plastic expansion&   tabulist&        max phenotype length   \\
GE & 0&           0&               9  &             11    &          69\\      
\end{tabular}
\end{table*}

\begin{table*}[!t]
\renewcommand{\arraystretch}{1.3}
\caption{The mean and {\it standard deviation} (in brackets) of the lengths of phenotypes collated from all 100 replications on each experimental setup (i.e., approximately 1 billion individuals sampled per setup). We observe larger phenotype lengths on average for the more successful setups.}
\label{phenotypelengthstable}
\centering
\footnotesize
\begin{tabular}{cccccc}
 &         &   asocial          &   asocial &     social   \\
 &  asocial&   plastic expansion&   tabulist&     max phenotype length   \\
GE &           2.68 ({\it 1.68})&      4.09 ({\it 3.35})&       5.16 ({\it 4.06})&       9.25 ({\it 13.70})\\      
\end{tabular}
\end{table*}

\section{Conclusion}
\label{conc}
We set out to explore the contribution that learning (both asocial and social) might play in the search for variable-length structures. The classic Hinton \& Nowlan~\cite{hinton&nowlan} problem landscape was adopted and experiments undertaken with grammar-based genetic programming operating under linear genome operators. A set of asocial and social learning mechanisms were examined with the result that we observe the setup that includes social learning to be the most successful. The setups exclusively adopting evolution with no lifetime learning are unable to solve the problem.
Future work will test the generalisation of these findings to a wider set of problem landscapes requiring the search for variable-length structures. Additionally, a broader set of learning mechanisms will be examined.

\end{document}